# Learning a binary search with a recurrent neural network. A novel approach to ordinal regression analysis

Louis Falissard (corresponding author):

CépiDc Inserm, Paris Saclay University, le Kremlin Bicêtre, France
Postal address: 31 rue du Général Leclerc, 94270, Le Kremlin Bicêtre, France
Email address: louis.falissard@inserm.fr
Phone number: +33679649178

Karim Bounebache : CépiDc, Inserm, Le Kremlin Bicêtre, France

Grégoire Rey : CépiDc, Inserm, Le Kremlin Bicêtre, France

Abstract: Deep neural networks are a family of computational models that are naturally suited to the analysis of hierarchical data such as, for instance, sequential data with the use of recurrent neural networks. In the other hand, ordinal regression is a well-known predictive modelling problem used in fields as diverse as psychometry to deep neural network based voice modelling. Their specificity lies in the properties of their outcome variable, typically considered as a categorical variable with natural ordering properties, typically allowing comparisons between different states ("a little" is less than "somewhat" which is itself less than "a lot", with transitivity allowed). This article investigates the application of sequence-to-sequence learning methods provided by the deep learning framework in ordinal regression, by formulating the ordinal regression problem as a sequential binary search. A method for visualizing the model's explanatory variables according to the ordinal target variable is proposed, that bears some similarities to linear discriminant analysis. The method is compared to traditional ordinal regression methods on a number of benchmark dataset, and is shown to have comparable or significantly better predictive power.

# 1   Introduction

Ordinal regression is a well-known predictive modelling problem used in fields as diverse as psychometry[1] to deep neural network based voice modelling[2]. Their specificity lies in the properties of their outcome variable, typically considered as a categorical variable with natural ordering properties, typically allowing comparisons between different states[3] ("a little" is less than "somewhat" which is itself less than "a lot", with transitivity allowed). Although this additional prior knowledge should be incorporated into the modelling process, this state comparability property ends up being surprisingly hard to integrate to pre-existing qualitative or quantitative approaches.

Indeed, most traditional ordinal regression methods (typically based on thresholding or least-square like modelling objectives) make additional assumptions on the outcome variable that might not always be verified[4]. As an example, the ordered logits model relies on the proportional odds assumption and the hypothesis that the observed ordered dependent variable constitutes an imperfect observation of a latent quantitative variable[5].

In computer science, the manipulation of ordered table is a well-known problem for which simple yet powerful algorithms have been known for decades. Binary search, for instance, allows for the localization of a given value in an ordered table of predefined size N using at most $\text{Log}_2(N)$ comparisons[6].

The following article proposes to make use of this simple algorithm's essence by encoding an ordered variable as a binary tree. The resulting modelling problem is then shown to be reminiscent of sequential models traditionally seen in the deep learning academic literature[7], and a recurrent neural network variant, the Gated Recurrent Unit[8] (GRU) is proposed to solve it. The predictive power of the investigated method is then assessed on a dozen openly available benchmark datasets. Comparison with traditional methods show a significant improvement in predictive power on a number of datasets, in term of both average error rate, squared Cohen Kappa score and squared error metrics.

## 2 Method

### 2.1 Ordinal variable encoding on a probabilistic binary search tree

A binary search is as simple yet powerful recursive algorithm that, from a sorted array, determines the position of one of its given element, by computing a logarithmic amount of comparison[6] between the investigated value and the table's elements as can be seen in figure 1:

- Select the median element of the table, and compare it to the investigated value
- If the median element is bigger than the investigated value, repeat the algorithm applied to the table's lower half
- If the median element is lower than the investigated value, repeat the algorithm applied to the table's bigger half
- If the median element is equal to the investigated value, stop the algorithm and return the median element's position

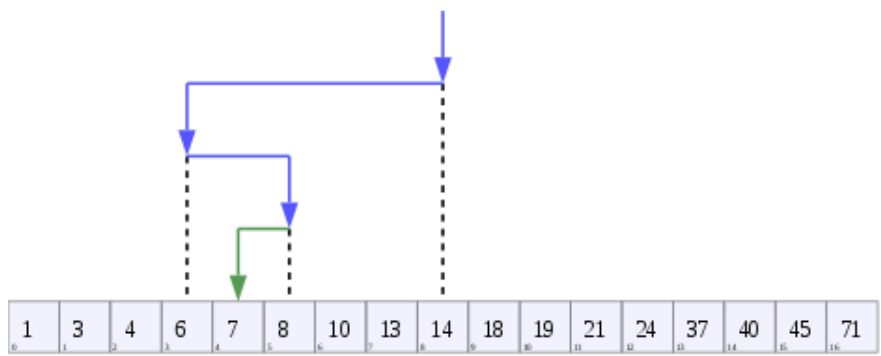

*Fig. 1 Example of a binary search algorithm (source: Wikipedia). The algorithm needs only 4 comparisons to find the position of an element in a table of 16 elements*

As a powerful approach to ordered sets manipulation relying solely on comparison operations, which are by definition perfectly acceptable in ordinal variable analysis, the binary search algorithm might constitute an interesting basis for the design of an ordinal regression method. For instance, directly applying the algorithm to an ordinal variable allows for its encoding on a binary tree. Figure 2 shows an example of such a decomposition, with an 8 states ordinal variable. Each path on the tree corresponds to a sequence of binary random variables defined as comparisons.

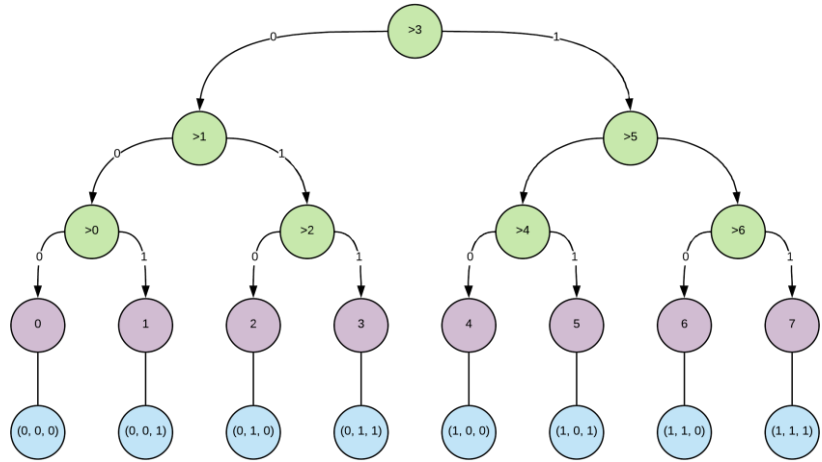

*Fig. 2 Binary search tree. Each state of the ordinal variable is associated with a vector of binary variables representing its location on the tree, and the result of the equivalent binary search process. Note that binary vectors correspond exactly to the binary decomposition of the ordinal variable's state*

By considering this binary search tree as a standard conditional tree diagram, and identifying the decision path leading to a given value corresponds to its decomposition in binary, ordinal regression can be formulated as a sequential modelling problem of binary variables as follows:

$$P(y \mid X, \theta) = P\left(\left\{\left\lfloor \frac{y}{2^{n-i}} \right\rfloor \bmod 2, i \in [\![1, n]\!]\right\} \mid X, \theta\right)$$

$$= \prod_{i=1}^{n} P\left(\left\lfloor \frac{y}{2^{n-i}} \right\rfloor \bmod 2 \mid \left\{\left\lfloor \frac{y}{2^{n-j}} \right\rfloor \bmod 2, j \in [\![1, i]\!]\right\}, X, \theta\right)$$

$$= \prod_{i=1}^{n} P(B_{i,n} \mid \{B_{j,n}, j \in [\![1, i]\!]\}, X, \theta) \text{ with } B_{i,n} = \left\lfloor \frac{y}{2^{n-i}} \right\rfloor \bmod 2 \; \forall (i, n) \in \mathbb{N}^2$$

Where:
- $y \in \mathbb{N}$ the ordinal dependent variable with $2^n, n \in \mathbb{N}$ states

- $X \in \mathbb{R}^d$, $\forall d \in \mathbb{N}$, the explanatory variables (in vectorial form)
- $\theta \in \mathbb{R}^e$, $\forall e \in \mathbb{N}$, the model's parameters

Note that so far, and for simplicity, the modelling problem is only defined for ordinal variable with a number of states that is a power of two (in other words where the ordinal variable's corresponding binary search graph is full). Extending the model to the general case of any given number of state is however quite straightforward and can be achieved as follows (and is shown in figure 3):

- Model the $n \in \mathbb{N}$ states ordinal variable as having $\left\lceil 2^{log_2(n)} \right\rceil$ states
- Force all the unnecessary states to 0 after model inference and renormalize the resulting probability distribution

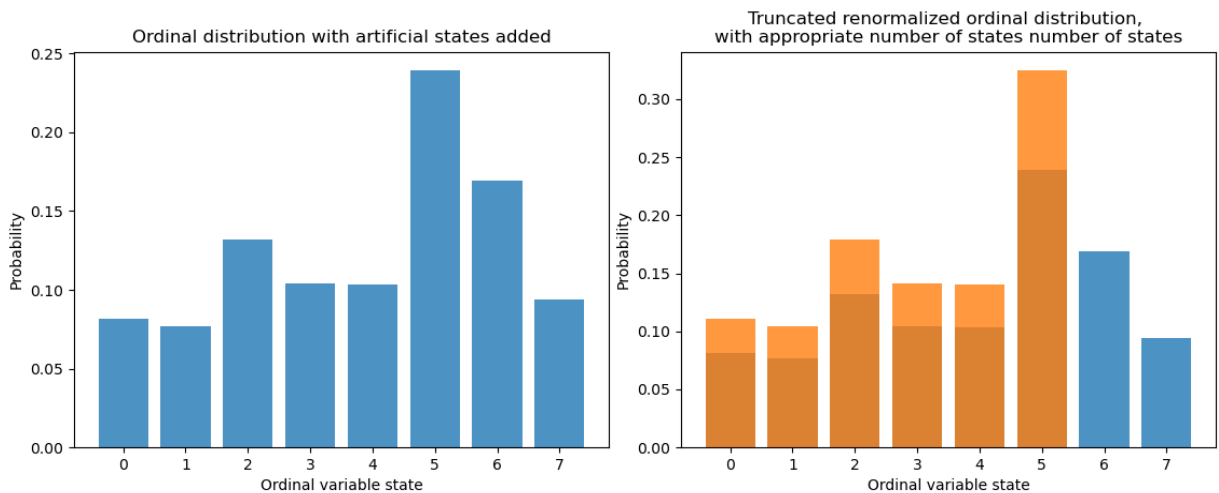

*Fig. 3: Example of truncation and renormalization trick for an ordinal variable with 6 different states. The variable is projected into a binary search tree of depth 3 (on the right). The resulting probability distribution is then modified such that the two higher states are associated to 0 probability (on the left)*

This type of modelling problem is highly reminiscent of sequential learning problems often seen in the deep artificial neural network academic literature[7]. Conditional language models, for instance, are derived by modelling sensibly similar sets of random variable with a shared dimensionality conditioned to each other in a sequential manner. Consequently, it only feels quite natural to investigate the potential of these methods to the aforedefined modelling problem, which this article proposes to solve using a very simple GRU-cell based seq2seq architecture. The authors are aware that this solution is now long considered out of fashion in the natural language processing community, but feel that the additional layers of complexity that came with modern neural translation architectures (such as attention modules) are too specific to their fields. They might however be the subject of future work.

## 2.2 Neural sequence models

In practical applications, neural sequence models are essentially used in natural language processing, specifically in machine translation where they have been representing the state of the art for a number of years[9]. However, in a broader sense, they provide machine learning practitioner with a powerful set of tools for the modelling of sequential, interdependent outcome variables. The first powerful neural sequence models were based on recurrent neural networks and their variants, such as Long Short Term Memory units[10] and Gated Recurrent Units. Although the deep learning academic community have found empirical evidence that now discourages their use in natural language processing tasks in favour of more modern approaches (attention[9] or dilated convolutions[11], for instance) for the sake of simplicity this approach was selected in order to model the aforedefined sequential decision modelling problem.

### 2.2.1 Recurrent neural networks

Recurrent neural networks are a family of neural network that specialize in the analysis of sequential data[12]. The main idea behind the elaboration of a recurrent neural network is to devise a model that shares its parameter across all time steps within the data sequence. Instead of feeding the whole sequence to a standard perceptron, each time step in the data is sequentially fed to the network, which also takes as input its previous output in order to allow the model to condition both on the present and past observations as can be seen on figure 4. As a recurrent neural network requires this past connection for each time steps, an additional input is given to the model when evaluating the first sequential observation. This vectorial input is called an initial state and is typically either set to 0 or considered as learnable parameters for the model[13].

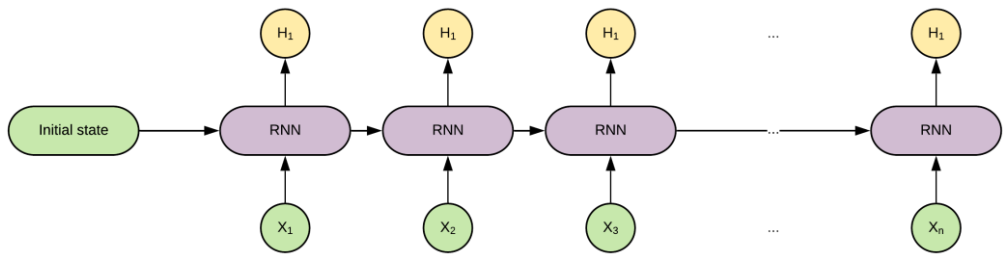

*Fig. 4: A recurrent neural network architecture outputs a vector for each sequential observation that depends on both the current and all previously observed time-steps*

This family of neural network can be used in a variety of settings that can be broadly gathered into 3 main categories that can be seen on figure 5:

- Modelling a non-sequential response variable from sequential explanatory variables (eg. Text classification)
- Modelling a sequential response variable from sequential explanatory variables (eg. Optical character recognition)
- Autoregressive modelling of a sequential variable (eg. Language models)

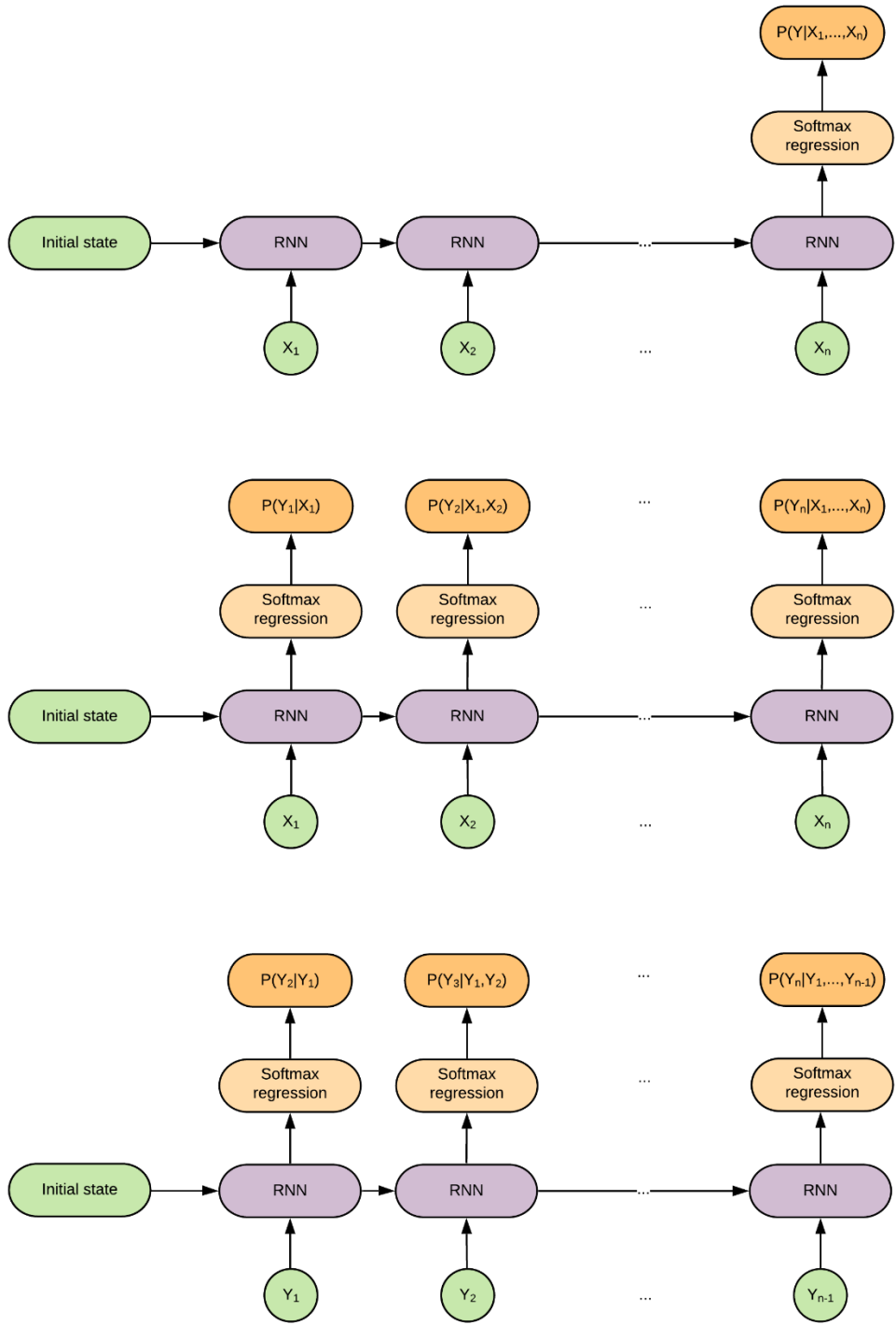

Fig. 5: Top: Recurrent neural network in regression setting. The RNN sweeps the entire input sequence, and its last output is used as inputs in a regression model conditioned on the entire sequence. Middle: RNN in sequential regression setting. The RNN sweeps the entire input sequence, and all of its outputs are used to fit a sequence of regression models conditioned on all previous observations. Bottom: RNN in autoregressive setting. The model sweeps the entire input sequence, and all of its outputs are used to fit a regression model to predict the next input from all previous observations

### 2.2.2 GRU

In order to extend traditional feedforward neural networks with this recurrent connections, several functional families have been devised over the years. Nowadays, the simplest, and original recurrent neural network is however typically discarded due to its poor behavior during model fitting[14], in favor of more modern approaches such as the Long Short Term Memory cell (LSTM) or the Gated Recurrent Unit (GRU). Although the LSTM cell has been shown to have a better modelling capability than the GRU, the latter was selected in the proposed approach for both its lesser amount of parameter and ability to handle smaller datasets. For a sequence of real valued vectorial input $(x_1, \dots, x_n)$ the output $h_t$ at time step t of a GRU is defined from both $h_{t-1}, x_t$ as follows[8]:

$$z_t = \sigma(W_z x_t + U_z h_{t-1} + b_z)$$

$$r_t = \sigma(W_r x_t + U_r h_{t-1} + b_r)$$

$$\hat{h}_t = \phi_h(W_h x_t + U_h(r_t \odot h_{t-1}) + b_h)$$

$$h_t = (1 - z_t) \odot \hat{h}_t + z_t \odot h_{t-1}$$

With:

- $x_t$ the input vector
- $h_t$ the output vector
- $\hat{h}_t$ the candidate activation vector
- $z_t$ the update gate vector
- $r_t$ the reset gate vector
- $\{W_i, U_i, b_i \; \forall \; i \in (z, r, h\,)\}$ learnable parameter matrices and vectors

### 2.2.3 Model definition

Amongst the basic recurrent neural network based architecture described in 2.2.1, the autoregressive setting seems like a good candidate to model the joint probability of observing a sequence of event. Indeed, sequentially modelling all the output variables conditioned on all previous ones almost allows for the computation of the joint modelling by simple product of all derived probabilities. However, a problem arises with simple autoregressive models that prevents their use as is in the investigated modelling problem. Autoregressive models expect the first sequence element in the as a given, which is not the case in the investigated modelling problem, where $P(B_{0,n})$ requires an estimate as well. The neural machine translation literature, which encounters the same problem, introduced the idea of adding a "neutral" state to the categorical variable (typically denoted as "<START>" in the machine translation academic literature) from which the network starts its auto-regression process[7]. This additional value is given to the recurrent network as its first input element, from which the model learns to predict the sequence's actual first element, as can be seen in figure 6.

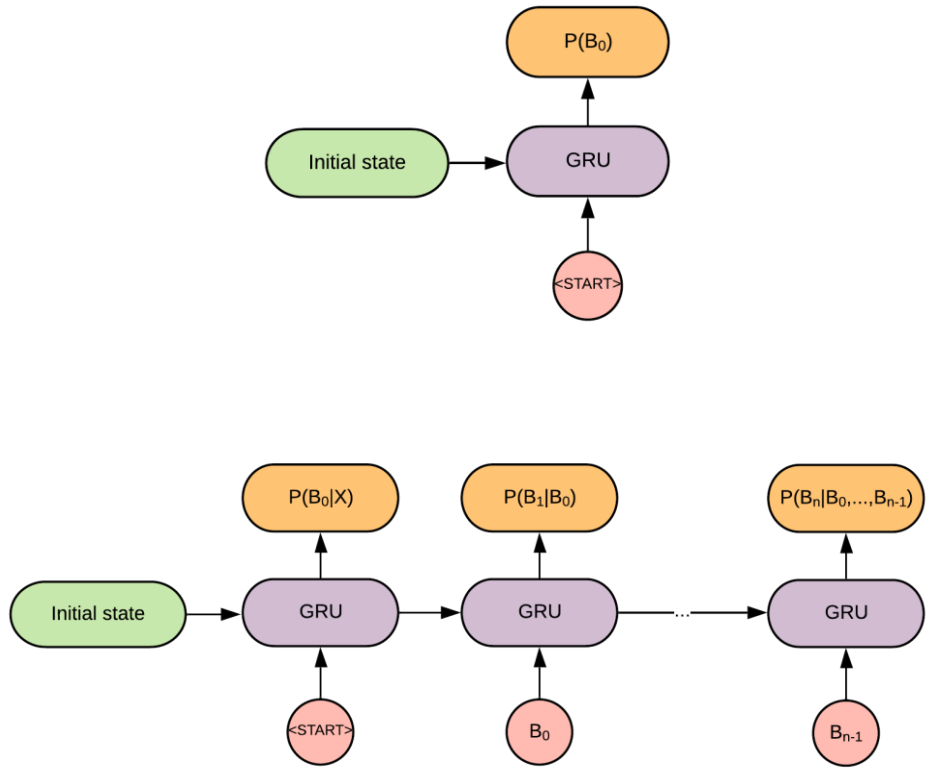

*Fig. 6 Left: A GRU recurrent neural network during the first step of the joint modelling autoregressive process. Its input is an additional, artificial state given to the target variable that never changes from one individual to another, that is used to estimate target sequence's first element's discrete probability density. Right: After the first recurrent neural network iteration, the actual sequence is given to the model, apart from the last element, which is never conditioned upon*

Although this neural architecture allows for efficient sequence joint probability modelling, it isn't by itself sufficient in order to solve the modelling problem investigated in this article, which, as defined in 2.1, consists in estimating the joint probability of binary decision sequence *conditioned* on some explanatory variables. The machine translation literature academic also had the same problem (eg. estimating the probability of a sentence in French given a sentence in English) and came up with several solutions. The simplest, found in early RNN based encoder-decoder architectures, was to make the recurrent neural network's initial state a function of the input variables, as can be seen in figure 7 which is the solution that was chosen for the here defined architecture.

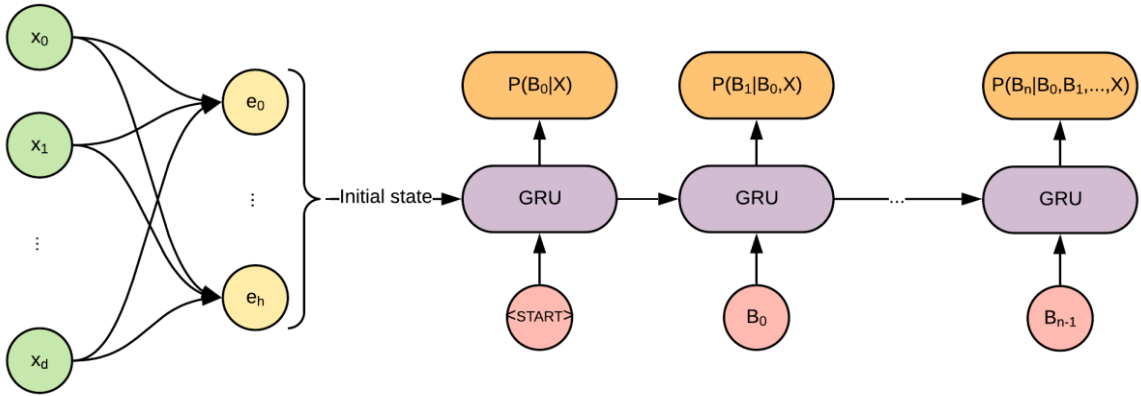

*Figure 7: A GRU neural network able to estimate the joint probability of a sequential output variable conditioned on some explanatory variables. h linear combinations of the explanatory variables are used as the initial state of a GRU that sweeps through the padded target sequence in an autoregressive fashion*

In summary, for an ordinal variable $Y$ with $2^n, n \in \mathbb{N}$ and explanatory variables $X \in \mathbb{R}^d$, $\forall d \in \mathbb{N}$, the entire model can be defined as follows, and its schematic representation can be seen in figure 8:

- The aforedefined random variables $B_{i,n}, i \in [\![1, n]\!]$ are encoded as two valued one hot vectors ($(0, 1)$ for $B_{i,n} = 0$, $(1, 0)$ for $B_{i,n} = 1$)
- The neutral state used to start the autoregressive token is defined as $(0, 0)$
- A GRU recurrent neural network with dimensionality $h$ is defined to sweep through the binary variable sequences (padded with the neutral state). $h$ constitutes the model's only hyper parameter (the authors advise to set this value to $n$, although without any theoretical nor empirical evidence to back it up)
- $h$ linear combination of the input variables are defined to build a vector $E$ that is to be used as the GRU's initial state

- A logistic regression is then applied to all of the GRU's outputs (the same logistic regression is applied at each time-step) in order to estimate the probability of each $B_{i,n}, i \in [\![1, n]\!]$ in an autoregressive fashion
- By feeding all possible binary decomposition sequences to the GRU (for the same individual), the ordinal variable probability can be retrieved from the logistic regression output's products
- The entire model (logistic regression, GRU and linear combination parameters) is jointly fit though maximum likelihood with gradient descent and backpropagation

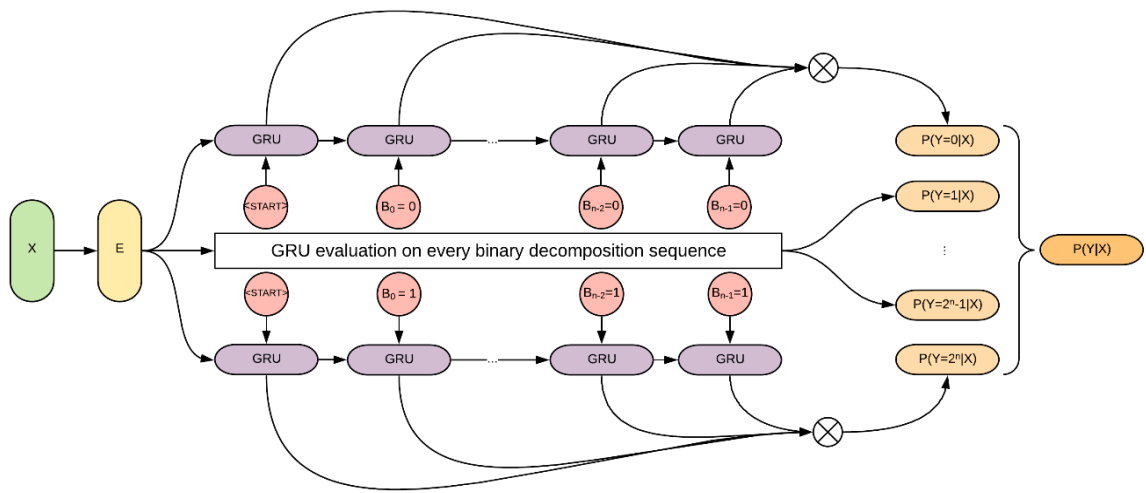

*Figure 8: The proposed neural architecture. From the explanatory variable, h linear combination are used to initialize a GRU that sweeps through all of the ordinal variable state's binary decompositions. The latter's actual distribution is then retrieved through product of all of the GRU's outputs*

### 2.2.4 Teacher forcing

The necessity of evaluating the recurrent neural network on every possible sequence in order to build the final probability distribution can result in significant computational needs, especially during model fitting, where model inference and gradient computation through backpropagation is required at each gradient descent iteration step. In order to speed up computation times, neural translation model are typically trained nowadays using a technique called teacher forcing[15]. Instead of fitting the model through maximum likelihood on the final joint distribution, model parameters are inferred by maximum likelihood on the sequential variables, and only the correct sequence is given to the model for each observation, as can be seen in figure 9. As a consequence, each optimization step requires the recurrent neural network to only assess one sequence per observation, thus significantly improving computation times.

All experiments reported in this article were derived using this model fitting approach. The final predictions used for performance estimations were however derived from the more traditional approach of building the entire ordinal probability distribution.

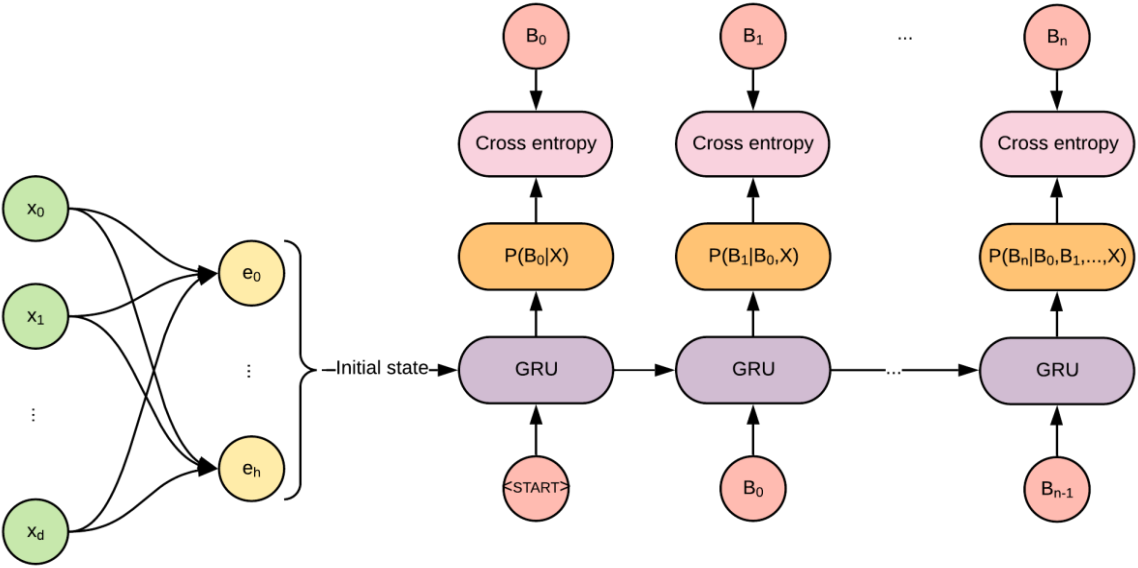

Figure 9: Example of model training using teacher forcing. The GRU's initial state is derived from linear combination of the input variables as before. However, to improve computation time, only the correct output sequence is fed to the recurrent neural network. The model is then jointly fit through maximum likelihood on each individual $B_{i,n}, i \in [\![1, n]\!]$

## 2.3 Linear dimensionality reduction for visualization

The method proposed in this article was primarily intended as a purely discriminant model. However, its inherent architecture can be exploited (at least in a majority of use cases) to use it as a linear dimensionality reduction and visualization tool as well, with no additional work required. Indeed, the GRU-cell's initial state, a vector based on linear combinations of the explanatory variable whose dimensionality constitutes the model's only hyper parameter, contains all the information the trained model uses to predict the target ordinal variable. Consequently, as long as this initial state's dimensionality is set to a value lower than the number of explanatory variables, it constitutes a compressed representation of the explanatory variables built to preserve as much discriminative information regarding the target variable as possible. This approach shares some similarity (at least conceptually) with partial least square regression methods, but is here specific to ordinal valued target variables.

In addition, when setting the model's hyper-parameter to values 2 or 3, the subsequent linear projection of the explanatory variables can be plotted, which allows for the visualization of potentially insightful patterns regarding the relationship between explanatory variables and the ordinal target.

In short, these linear projections of the explanatory variables can be obtained as follows:

- Set the model's hyper-parameter to a value lower than the explanatory variables cardinality
- Train the model
- Discard the GRU cell from the model and only compute the linear projections used to build its initial state

Finally, as this projection is purely linear, the authors have some hope that they can retain some interpretability, which is quite rare in neural network models. However, additional work is required to properly assess how these linear combinations can be interpreted.

# 3 Experiment

## 3.1 Description

To assess its predictive performances, the aforementioned method was applied on a set of readily available benchmark datasets for ordinal prediction, and compared to results obtained from more traditional approaches. All the ordinal regression methods used in the following experiment (except for the one introduced in this article) were taken from the mord Python package, and all roughly follow two different approaches:

- Threshold based methods, comprised of three variants of the ordinal logistic model, the all-threshold ordinal logistic model, the immediate-threshold ordinal logistic model and the squared error ordinal logistic model, which are referred to as "AT", "IT" and "SE" in the experiment's results
- The regression based methods, comprised of the ordinal ridge regression model, and the least absolute deviation ordinal regression model, which are referred to as "Ridge" and "LAD" in the experiment's results

| Dataset | Number of ordinal states in the output variable | Number of input variables | Sample siwe |
|---|---|---|---|
| Abalone | 8 | 7 | 4177 |
| Abalone_ord | 10 | 10 | 4177 |
| Affairs | 6 | 17 | 265 |
| Ailerons | 9 | 39 | 7154 |
| Auto_ord | 10 | 8 | 398 |
| Auto_riskness | 6 | 15 | 160 |
| Bostonhousing_ord | 5 | 13 | 506 |
| Boston_housing | 6 | 13 | 506 |
| California_housing | 6 | 8 | 20640 |
| Cement_strength | 5 | 8 | 998 |
| Fireman_example | 16 | 10 | 40768 |
| Glass | 6 | 9 | 213 |
| Kinematics | 8 | 8 | 8192 |
| Machine_ord | 10 | 6 | 199 |
| Skill | 7 | 18 | 3337 |
| Stock_ord | 5 | 9 | 950 |
| Winequality_red | 6 | 11 | 1359 |
| Winequality_white | 7 | 11 | 3961 |

*Table 1: Datasets summary*

The "Wisconsin_breast_ord" dataset, also readily available in the same source for ordinal regression method benchmarking, was discarded for the experiments due to its low observation to sample size ratio (only 194 observations for 33 variables). Indeed, as it stands now, the ordinal regression method presented in this dataset is not meant as a tool for scarce datasets as Ridge or Lasso regressions are. However, the application of such penalty based regularization methods will be the object of future work.

In order to assess the proposed method's performances in comparison to the state of the art, all these methods were used on 17 real-life datasets traditionally used for benchmarking ordinal regression methods. A summary of these datasets can be found in table 1. The following methodology was used for the experiment:

- Each input variable in all dataset was standardized to zero mean and unitary standard variance
- Every model (the proposed approach, and the mord package's model) was fit to each dataset and three performance metrics were assessed using 10-fold cross validation: Model accuracy, squared cohen kappa score and squared error. Confidence intervals were estimated through bootstrap

An additional experiment was designed in order to assess the model's visualization capability. A supplementary model with hyper-parameter value set to two was adjusted to each of the aforedescribed dataset. The resulting bi-dimensional embedding were then plotted against the ordinal target value in order to qualitatively assess whether these linear projections can indeed capture interesting patterns in the data.

## 3.2 Results

### 3.2.1 Predictive performances

The predictive performance experiment's results are displayed in tables 2, 3 and 4 for accuracy, squared Cohen Kappa and mean squared error metrics respectively. For readability, only the best baseline method's performance metrics are reported for each dataset, and scores showing a significantly better performance are highlighted in bold. The interested reader can however find the experiment's complete results in the annex. For 7 datasets (namely "Auto_riskness", "Boston_housing", "Bostonhousing_ord", "Glass", "Machine_ord", "Skill" and "Winequality_red"), no significant difference in predictive performance could be find between the proposed approach and the best baseline method in all investigated metrics. For the 10 remaining datasets, significant differences in predictive power were found, and can be summed up as follows:

- For four datasets, namely "Cement_strength", "Fireman_Example", "Kinematics" and "Stock_ord"), the proposed approach significantly outperformed the best baseline method on all metrics

- For one dataset, namely "Affairs", the proposed approach was significantly outperformed by at least one baseline method on all assessed metrics. However, the best baseline approach for this dataset differs for all metrics (Logistic IT for accuracy, logistic AT for Cohen Kappa, and Ridge for the mean squared error)

- When focusing only on accuracy, the proposed approach outperforms all baseline methods on an additional three datasets ("Abalone", "Abalone_ord" and

"California_housing"), and isn't outperformed by any of them on any additional dataset beside "Affairs"

- When focusing only on squared Cohen Kappa score, the proposed approach outperforms all baseline methods on an additional two datasets ("California_housing" and "Winequality_white"), and isn't outperformed by any of them on any additional dataset beside "Affairs"

- When focusing only on mean squared error, the proposed approach does not significantly outperforms all baseline methods on any additional dataset. It is however outperformed by an additional dataset, namely "Abalone"

| Dataset | Proposed | Best other | Method |
|---|---|---|---|
| abalone | **37.6 [36.1, 39.1]** | 32.8 [31.4, 34.1] | IT |
| abalone_ord | **58.7 [57.2, 60.1]** | 55.2 [53.7, 56.7] | LAD |
| affairs | 25.0 [20.0, 30.4] | **48.1 [41.9, 54.2]** | IT |
| ailerons | 45.5 [44.3, 46.7] | 43.2 [42.0, 44.3] | IT |
| auto_ord | 51.0 [46.2, 55.9] | 55.4 [50.5, 60.3] | SE |
| auto_riskness | 66.9 [59.4, 73.8] | 63.1 [55.6, 70.0] | LAD |
| bostonhousing_ord | 73.6 [69.6, 77.4] | 72.0 [68.0, 75.8] | AT |
| boston_housing | 56.6 [52.2, 61.0] | 61.6 [57.4, 65.8] | IT |
| california_housing | **57.9 [57.3, 58.6]** | 54.0 [53.3, 54.7] | AT |
| **cement_strength** | **69.1 [66.2, 71.9]** | 49.4 [46.2, 52.5] | LAD |
| **fireman_example** | **39.9 [39.4, 40.4]** | 23.0 [22.6, 23.5] | IT |
| glass | 57.1 [50.5, 63.8] | 56.2 [49.5, 62.9] | IT |
| **kinematics** | **42.7 [41.6, 43.8]** | 27.4 [26.5, 28.4] | IT |
| machine_ord | 57.4 [50.5, 64.2] | 66.3 [59.5, 73.2] | AT |
| skill | 41.6 [39.9, 43.3] | 40.5 [38.7, 42.1] | AT |
| **stock_ord** | **85.7 [83.5, 87.8]** | 69.8 [66.8, 72.6] | IT |
| winequality_red | 57.9 [55.2, 60.6] | 58.0 [55.3, 60.6] | LAD |
| winequality_white | 53.9 [52.4, 55.5] | 52.9 [51.3, 54.4] | LAD |

*Table 2: Accuracy results (in %)*

| Dataset | Proposed | Best other | Method |
|---|---|---|---|
| abalone | 74.6 [73.0, 76.1] | 72.9 [71.4, 74.4] | AT |
| abalone_ord | 62.5 [60.5, 64.5] | 63.6 [61.6, 65.5] | Ridge |
| affairs | -11.1 [-23.0, 1.3] | **23.0 [12.2, 33.7]** | AT |
| ailerons | 89.5 [89.0, 90.1] | 89.6 [89.1, 90.1] | AT |
| auto_ord | 88.5 [86.0, 90.6] | 91.1 [89.3, 92.6] | SE |
| auto_riskness | 61.3 [44.2, 75.6] | 66.2 [56.3, 74.7] | AT |
| bostonhousing_ord | 82.3 [77.6, 86.3] | 82.5 [78.4, 85.9] | SE |
| boston_housing | 85.9 [82.7, 88.6] | 87.3 [84.4, 90.0] | IT |
| california_housing | **79.1 [78.4, 79.8]** | 77.7 [77.0, 78.4] | AT |
| **cement_strength** | **88.0 [86.4, 89.5]** | 71.3 [68.0, 74.2] | IT |
| **fireman_example** | **96.3 [96.2, 96.4]** | 84.3 [84.0, 84.7] | AT |
| glass | 71.3 [60.5, 80.1] | 80.4 [73.4, 85.7] | LAD |
| **kinematics** | **84.5 [83.7, 85.3]** | 62.0 [60.6, 63.4] | IT |
| machine_ord | 80.7 [67.8, 89.9] | 92.4 [87.0, 95.7] | SE |
| skill | 73.0 [71.4, 74.6] | 70.6 [68.9, 72.2] | IT |
| **stock_ord** | **95.1 [94.2, 95.9]** | 88.6 [87.0, 90.1] | IT |
| winequality_red | 50.4 [46.1, 54.5] | 49.5 [45.8, 53.2] | LAD |
| winequality_white | **49.0 [46.8, 51.2]** | 43.2 [40.9, 45.6] | LAD |

*Table 3: Quadratic Cohen Kappa results (in %)*

| | Proposed | Best other | Method |
|---|---|---|---|
| abalone | 2.46 [2.33, 2.59] | **2.13 [2.03, 2.22]** | SE |
| abalone_ord | 0.77 [0.72, 0.83] | 0.76 [0.71, 0.81] | Ridge |
| affairs | 7.90 [6.83, 9.01] | **3.45 [2.99, 3.93]** | Ridge |
| ailerons | 1.45 [1.38, 1.52] | 1.35 [1.30, 1.41] | SE |
| auto_ord | 0.98 [0.78, 1.20] | 0.73 [0.60, 0.87] | SE |
| auto_riskness | 1.14 [0.69, 1.71] | 0.81 [0.62, 1.03] | AT |
| bostonhousing_ord | 0.38 [0.29, 0.47] | 0.34 [0.28, 0.40] | SE |
| boston_housing | 0.73 [0.61, 0.88] | 0.64 [0.52, 0.78] | IT |
| california_housing | 0.76 [0.74, 0.78] | 0.77 [0.75, 0.79] | SE |
| **cement_strength** | **0.35 [0.31, 0.39]** | 0.74 [0.68, 0.81] | SE |
| **fireman_example** | **1.58 [1.55, 1.61]** | 6.09 [5.99, 6.19] | SE |
| glass | 1.67 [1.18, 2.24] | 1.01 [0.77, 1.29] | Ridge |
| **kinematics** | **1.66 [1.59, 1.74]** | 3.22 [3.13, 3.31] | SE |
| machine_ord | 1.92 [1.02, 3.18] | 0.79 [0.44, 1.25] | SE |
| skill | 1.02 [0.96, 1.07] | 1.03 [0.98, 1.08] | SE |
| **stock_ord** | **0.14 [0.12, 0.17]** | 0.32 [0.29, 0.36] | IT |
| winequality_red | 0.56 [0.51, 0.61] | 0.53 [0.48, 0.57] | LAD |
| winequality_white | 0.61 [0.58, 0.65] | 0.65 [0.62, 0.69] | LAD |

*Table 4: Mean squared error results*

In order to provide better insight on the proposed approach's performances against all baseline methods, figure 10 displays for each dataset which of the baseline methods either significantly outperform, is outperformed, or does not perform significantly differently than the proposed approach, for all chosen metrics.

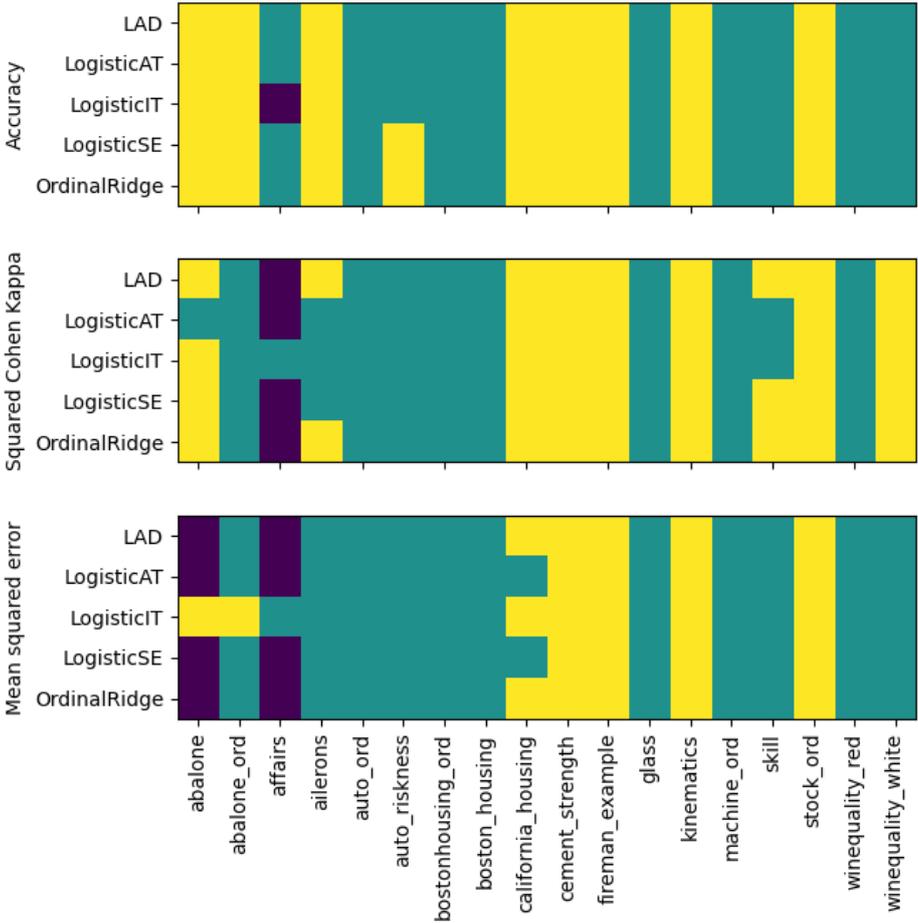

Figure 10: Results comparisons between the proposed approach and all baseline methods on all datasets. Yellow, blue and purple cases denote baseline methods that respectively perform significantly worse, not significantly better or worse and significantly better than the proposed approach on the given dataset

As can be seen on figure 10, the proposed approach significantly outperforms any baseline approach on any given dataset for a total number of 106 times, and is significantly outperformed 13 times.

Moreover, the binary search based method is never significantly beaten by all baseline methods on any of the investigated datasets, this for all chosen metrics.

As was already shown in table 4, the proposed approach's performances in term of mean squared error are a bit weaker. Indeed, it is outperformed by all baseline methods besides the "LogisticIT" on both the "Abalone" and the "Affairs" datasets. These poorer performances might be explained by the fact that the proposed method's approach does not rely on any mean squared (or mean squared surrogates) objective for model fitting. In any cases, additional analysis of these datasets to better understand these poorer performances will be treated in the discussion

### 3.2.2 Linear dimensionality reduction for ordinal visualization

As previously described, in order to assess the proposed approach' potential for data visualization, a set of additional models were fit to each dataset, with hyper-parameter set to 2 to allow for efficient scatterplot.

Some of the resulting bi-dimensional projections of the input data can be seen in figure X, with each point color-mapped according to its ordinal target variable value. The remaining visualizations can be found in the annex, with varying results. For instance, as the "affairs" dataset suffered from extremely poor prediction performances, it is reasonable to expect its resulting projection to yield few to no insight about the relationship between target and explanatory variables.

Insights gathered from the visualizations displayed in figure 11 can be summed up in two major points:

- When it comes to the "Fireman example" and "Ailerons" datasets, the relationship between the target variable and the explanatory variable linear combinations is quite smooth and progressive. The "Fireman example" dataset however appears to have non-linear decision boundaries

- For the "Boustonhousing ord" and "Stock ord" datasets, however, the relationship between explanatory and target variables is not as straightforward. Indeed, the visualizations show clusters of data points that each keep an ordered relationship with the target variable. However, the decisions boundaries are not the same for all clusters, indicating that stratification might be of interest in the analysis of these datasets.

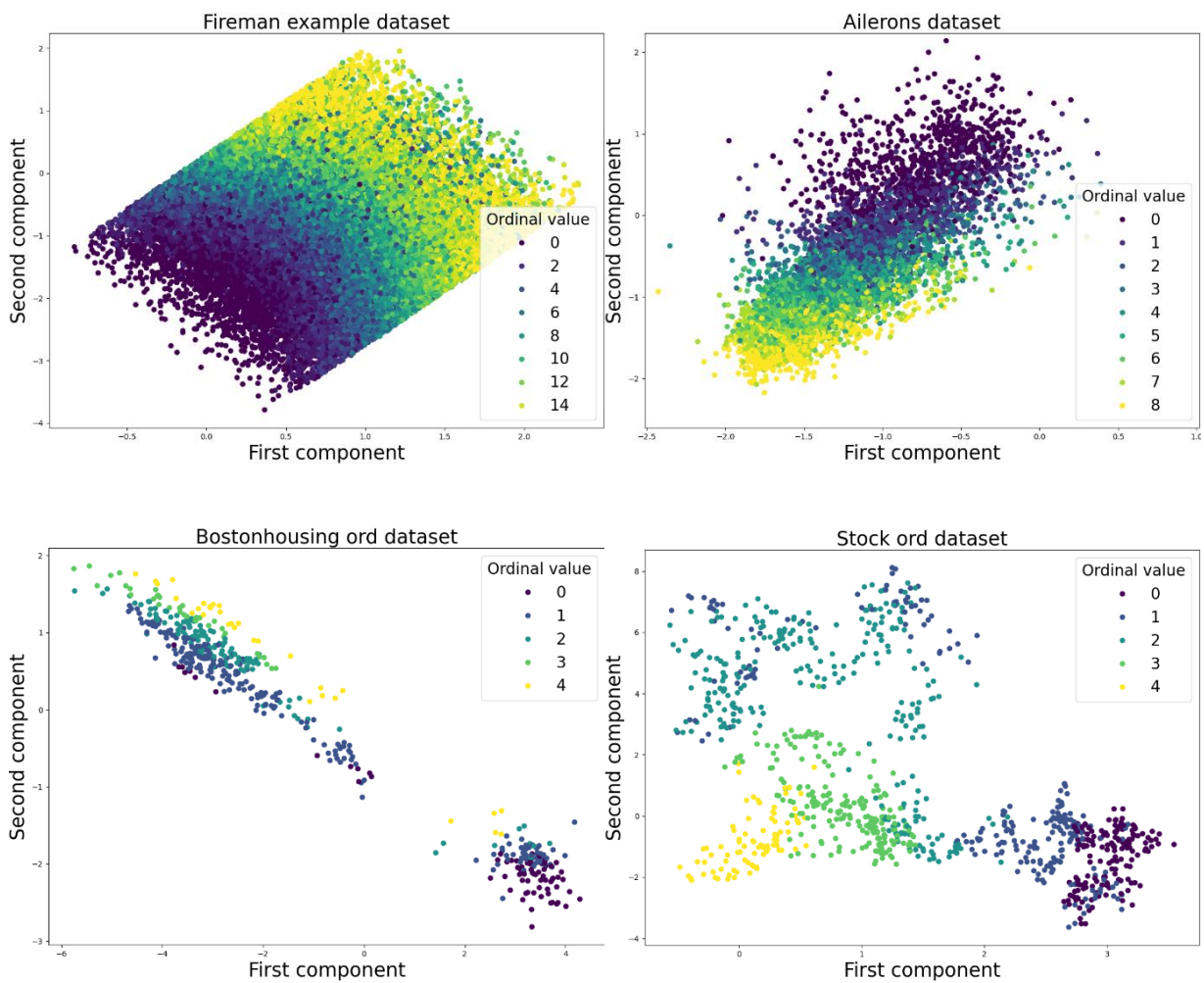

Fig. 11 Examples of bi-dimensional projections obtained by fitting the proposed model (with number of neurons in the recurrent network parameter fixed to 2) to a selected sample of datasets used in the experiment. For each dataset, the linear projections lead to highly readable visualization of the explanatory variables' relationship to the ordinal target variable

# 4 Discussion

As previously seen in the result part, the proposed approach tends to yield better or similar predictive performances to all baseline approach on most datasets, with the exception of the « affairs » and « abalone » datasets. Consequently, developing a better understanding of these datasets might be helpful in order to assess cases where the approach for ordinal regression presented in this article might not be advisable. In addition, it might also lead to empirically derived conjectures regarding hypotheses the model requires in order to perform well, or provide elements that might lead to further improvements.

## 4.1 "Affairs" dataset

The "affairs" dataset constitutes the dataset on which the proposed approach' overall performances are the lowers. Indeed, it is the only investigated dataset where recurrent neural network based ordinal regression is outperformed by at least one baseline method for all selected performance metrics.

The first thing that can be noticed about the "affairs" dataset is its low sample size compared to its high number of explanatory variables. Indeed, this dataset is comprised of 265 observations each comprised of 17 explanatory variables. In addition, its ordinal target variable is made of 6 different states. Such a poor dimensionality to sample ratio typically requires regularization methods. In addition, neural network based methods for predictive modelling are known to easily overfit. However, no regularization methods were used during model training in the experiments presented in this article, which might explain the model's poor performance. This hypothesis is further confirmed by assessing the model's differences in performance between the training and validation dataset, that are displayed in table 5. Indeed, a significant gap can be observed between training and validation metrics, and constitutes strong evidence indicating the model is overfitting the dataset.

| Dataset | Accuracy (%) | Cohen Kappa | Mean Squared Error |
| --- | --- | --- | --- |
| Training | 54 | .42 | 4.1 |
| Validation | 27 | -.53 | 7.9 |

Table 5: Training and validation performance metrics for the "affairs" dataset. The significant decrease in performance from training to validation suggests that the model is strongly overfitting the dataset

As a consequence, incorporating regularization methods in the proposed approach should be the object of future work. A promising candidate to do so can be found in the dropout method, traditionally used in deep learning models, which could be applied to the recurrent neural network part of the presented architecture. Another solution that might be of interest lies in penalized methods. Indeed, one could add a lasso, ridge or elastic-net penalization to the objective function. This penalization could typically be applied to the linear combination weights that are used to build the recurrent neural network's initial state. For a lasso penalization applied on these weights, for instance, the feature selection interpretation of lasso regression methods would remain heuristically valid.

Another remarkable property of the "affairs" dataset is that in addition to having considerably low sample size, it is considerably unbalanced. Indeed, as can be seen in figure 12, more than half of the dataset's observations are associated with a target value of 1, with some target values only observed as few as 15 times. Consequently, sampling or loss weighting techniques should be considered necessary in order to properly solve this modelling problem.

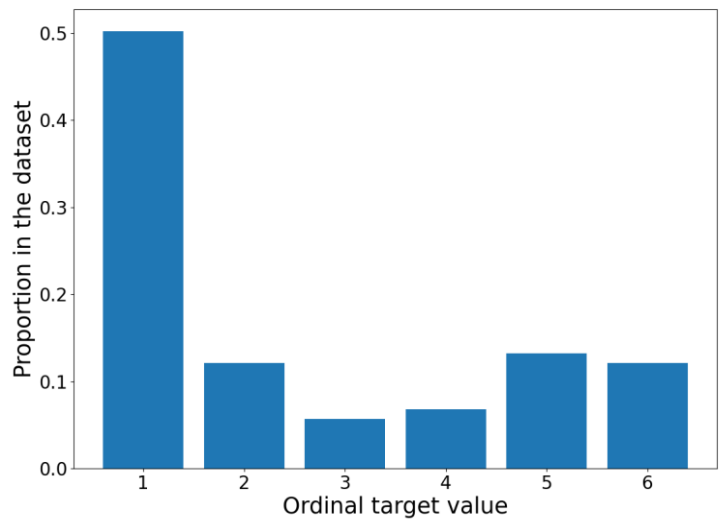

Fig. 12 Distribution of the ordinal target variable in the dataset. The dataset is extremely unbalanced. Approximately 50% of observations correspond to the first target value. All other value have less than 35 observations

Although sampling techniques can perfectly adapt to the proposed approach without any additional work, loss re-weighting techniques are not as straightforward, especially when training with teacher forcing. Indeed, in teacher forcing, the actual labels used to fit the model are the binary decomposition of the target values. As such, weighting methods should be adapted in order for weighting to apply for this sequence of target variable.

## 4.2 "Abalone" dataset

Although not as concerning in terms of predictive performance, compared to the "affairs" dataset, the model's behaviour on the "abalone" dataset is more complicated to explain. Indeed, the model is only significantly outperformed on mean squared error metrics, and significantly outperforms almost all baseline approaches on the two other selected metrics (apart from the "LogisticAT" model with regard to the quadratic Cohen Kappa score). However, the authors could not find any satisfactory explanation for this unique behaviour. Indeed, the dataset's dimensionality (7 explanatory variables for approximately 4 thousands data points) seems quite sufficient, which is further confirmed when estimating performance metrics on the training set, which are essentially identical to those evaluated through cross validation. In addition, the distribution of the ordinal target variable does not suffer from severe unbalance such as could be observed with the "affairs" dataset.

However, the visualization capability of the proposed model can constitute a way to further analyse this dataset in order to build hypotheses that might explain this phenomenon, at least qualitatively. Figure 13 shows the two dimensional projection of the "abalone" dataset's explanatory variables, with each point color coded in three different manners:

- Each point color coded according to the ordinal variable's true value
- Each point color coded according to the proposed approach' prediction
- Each point color coded according to the "LogisticSE"'s (best model in term of mean squared error) prediction

Qualitatively, the proposed approach seems to derive decision surfaces that better fit the true ordinal values than the "LogisticSE" does (which is further confirmed by its significantly better performance in terms of accuracy). However, these boundaries are not entirely ordinal. Although decision boundaries for states 0 through 6 are organized in a fairly sequential, ordered approach, state 7's decision surface appears to be adjacent to states 3, 4 and 6's boundaries. This adjacency property is lost when using the baseline models, that all have as a hypothesis either a mean squared property or a proportional odds assumption (guaranteeing parallel decision surfaces). As such, the baseline approaches on these dataset might be biased towards maximizing the mean squared error at the detriment of finding true decision surfaces.

As a consequence, one could emit the hypothesis that the ordinal properties of the "abalone" dataset's target variables are not as clear cut as would be expected, and that mean squared error metrics are not quite perfectly fit to evaluate model performances on this particular dataset.

To conclude, figure 13 shows a potential use of the visualization for qualitative exploration of datasets in relation to an ordinal variable. Indeed, it allowed us to propose an hypothesis to explain the model's behaviour on the "abalone" dataset by investigating the decision boundaries derived by different models. In addition, the reader can notice that the decision surfaces of the "LogisticSE" model are remarquably preserved by the projection, showing that these linear combination of the explanatory variables do capture efficient representations of the explanatory variables.

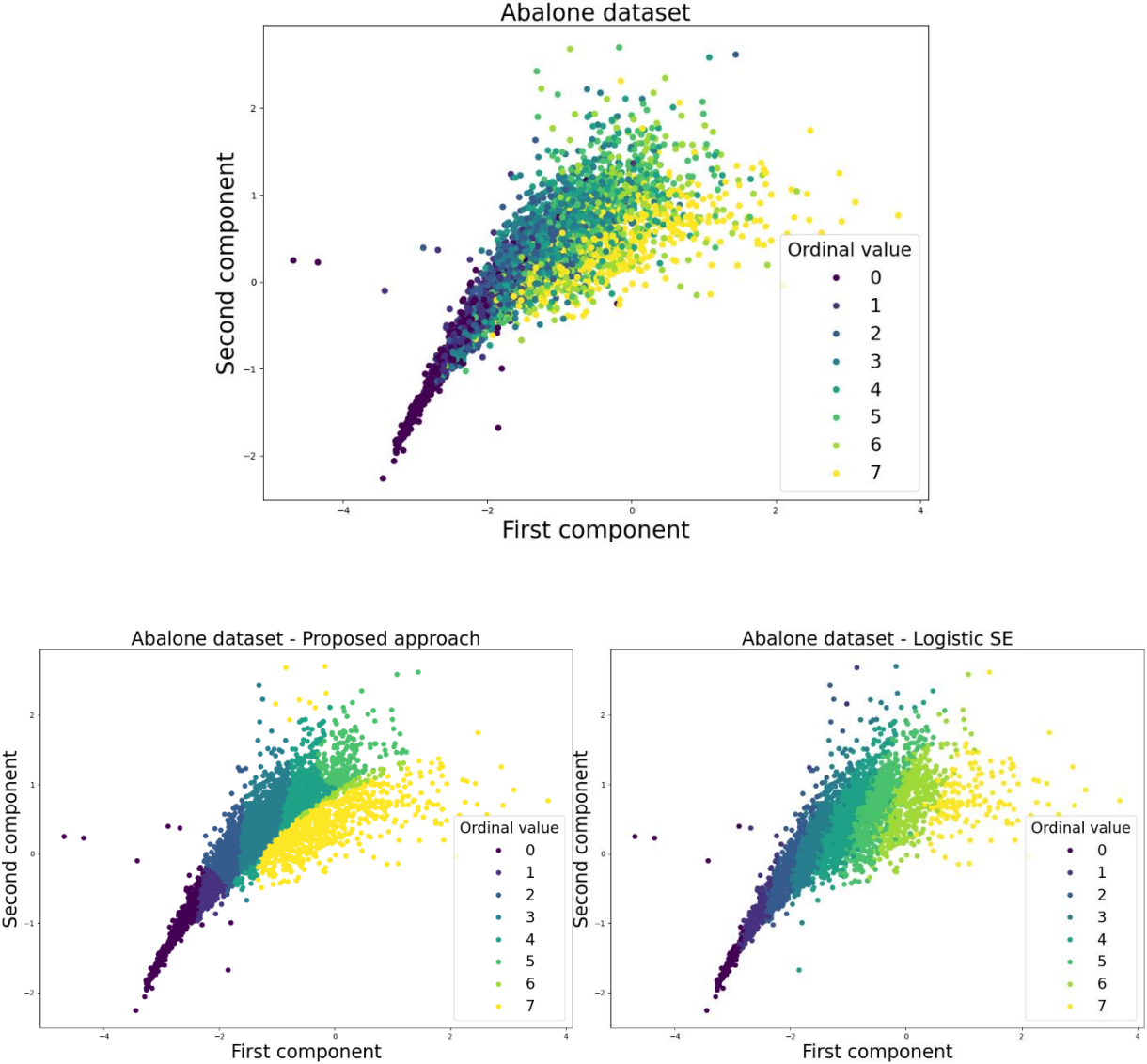

Figure 13: Top: Bi-dimensional projection of the input variables color-coded according to the ordinal target values. Bottom-left: Same projections color-coded according to the proposed approach' predictions of the target variable. Bottom-left: Same projections colo-coded according to the Logistic SE' predictions of the target variable.

# 5  Conclusion

In this article, we presented a formulation of the ordinal regression modelling problem as a sequential learning problem based on a binary search. We then proposed to solve said problem with a simple recurrent neural network. The derived approach was shown to outperform most traditional methods currently used for ordinal regression analysis on a number of datasets, this according to three different metrics, namely mean accuracy, quadratic Cohen Kappa score and mean squared error. Although the method was outperformed on two distinct datasets, reasonable hypothesis were found in order to explain the phenomenon. In addition, we showed that this method allows practitioners to build efficient, low dimensional representations of the investigated ordinal dataset, that can for instance be used to assess qualitatively how ordinal the investigated target variable actually is. These representations are in addition based on linear combinations of the explanatory variables, and as such share some similarity with discriminant analysis, at least conceptually. As a consequence, investigating whether the derived linear combinations' coefficients values have theoretical interpretation could be an interesting lead for future work, and might result in a rare case of (at least partially) interpretable deep learning based predictive model.

# 6  Bibliography

1. Turkoz, I., Fu, D.-J., Bossie, C. A., Sheehan, J. J. & Alphs, L. Relationship between the clinical global impression of severity for schizoaffective disorder scale and established mood scales for mania and depression. *J Affect Disord* **150**, 17–22 (2013).

2. Oord, A. van den *et al.* WaveNet: A Generative Model for Raw Audio. *arXiv:1609.03499 [cs]* (2016).

3. Winship, C. & Mare, R. D. Regression Models with Ordinal Variables. *American Sociological Review* **49**, 512 (1984).

4. Pedregosa-Izquierdo, F. Feature extraction and supervised learning on fMRI : from practice to theory. (Université Pierre et Marie Curie - Paris VI, 2015).

5. McCullagh, P. Regression Models for Ordinal Data. *Journal of the Royal Statistical Society: Series B (Methodological)* **42**, 109–127 (1980).

6. A modification to the half-interval search (binary search) method | Proceedings of the 14th annual Southeast regional conference. https://dl.acm.org/doi/10.1145/503561.503582.

7. Sutskever, I., Vinyals, O. & Le, Q. V. Sequence to Sequence Learning with Neural Networks. in *Advances in Neural Information Processing Systems 27* (eds. Ghahramani, Z., Welling, M., Cortes, C., Lawrence, N. D. & Weinberger, K. Q.) 3104–3112 (Curran Associates, Inc., 2014).

8. Cho, K. *et al.* Learning Phrase Representations using RNN Encoder-Decoder for Statistical Machine Translation. *arXiv:1406.1078 [cs, stat]* (2014).

9. Vaswani, A. *et al.* Attention Is All You Need. *arXiv:1706.03762 [cs]* (2017).

10. Hochreiter, S. & Schmidhuber, J. Long Short-Term Memory. *Neural Computation* **9**, 1735–1780 (1997).

11. Gehring, J., Auli, M., Grangier, D., Yarats, D. & Dauphin, Y. N. Convolutional Sequence to Sequence Learning. *arXiv:1705.03122 [cs]* (2017).

12. Mozer, M. A Focused Backpropagation Algorithm for Temporal Pattern Recognition. *Complex Systems* **3**, (1995).

13. Abiodun, O. I. *et al.* State-of-the-art in artificial neural network applications: A survey. *Heliyon* **4**, (2018).

14. Informatik, F., Bengio, Y., Frasconi, P. & Schmidhuber, J. Gradient Flow in Recurrent Nets: the Difficulty of Learning Long-Term Dependencies. *A Field Guide to Dynamical Recurrent Neural Networks* (2003).

15. LeCun, Y., Bengio, Y. & Hinton, G. Deep learning. *Nature* **521**, 436–444 (2015).